\def\paperlanguage{}
\pdfoutput=1 % for arxiv

\documentclass[letterpaper, 10 pt, conference]{ieeeconf}  % Comment this line out if you need a4paper

\usepackage{bm}
\usepackage{cite}
\usepackage{flushend}
\include{preamble}

\IEEEoverridecommandlockouts                              % This command is only needed if
\title{\LARGE \textbf
  {
    \switchlanguage%
    {%
      Dynamic Task Control Method of a Flexible Manipulator\\Using a Deep Recurrent Neural Network
    }%
    {%
      再帰型深層学習を用いた柔軟マニピュレータによる動的なタスク実行
    }%
  }
}

\author{Kento Kawaharazuka$^1$, Toru Ogawa$^2$, and Cota Nabeshima$^2$% <-this % stops a space
  \thanks{$^1$ An author is associated with Department of Mechano-Informatics, Graduate School of Information Science and Technology, The University of Tokyo. %, 7-3-1 Hongo, Bunkyo-ku, Tokyo, 113-8656, Japan.
    \texttt\small kawaharazuka@jsk.t.u-tokyo.ac.jp
  }
  \thanks{$^2$ Authors are associated with Preferred Networks, Inc. %
    \texttt\small \{ogawa, cota\}@preferred.jp
  }
  \thanks{
  This work is an achievement during part-time job at Preferred Networks. %
  }
}
\begin{document}

\maketitle
\thispagestyle{empty}
\pagestyle{empty}

%%%%%%%%%%%%%%%%%%%%%%%%%%%%%%%%%%%%%%%%%%%%%%%%%%%%%%%%%%%%%%%%%%%%%%%%%%%%%%%%
\begin{abstract}
  \switchlanguage%
  {%
    %While the flexible body has advantages over the rigid body in terms of motions with environmental contact due to its underactuation, its control is difficult.
    %So we need to develop a method for the flexible body to realize a certain task dynamically and precisely.
    %Therefore, we construct a versatile deep recurrent neural network which generically represents the transitions, due to control commands, of the flexible body state and task state, and calculate optimized control commands to realize the task using this network.
    %Although this study can be applied to various manipulators and tasks, as one example, we choose the task of Wadaiko (traditional Japanese drum) drumming by a flexible manipulator, which outputs only sparse events and is difficult for conventional methods to handle, and verify the effectiveness of this study by realizing the intended timing and volume of sound using our network.
    The flexible body has advantages over the rigid body in terms of environmental contact thanks to its underactuation.
    On the other hand, when applying conventional control methods to realize dynamic tasks with the flexible body, there are two difficulties: accurate modeling of the flexible body and the derivation of intermediate postures to achieve the tasks.
    Learning-based methods are considered to be more effective than accurate modeling, but they require explicit intermediate postures.
    To solve these two difficulties at the same time, we developed a real-time task control method with a deep recurrent neural network named Dynamic Task Execution Network (DTXNET), which acquires the relationship among the control command, robot state including image information, and task state.
    Once the network is trained, only the target event and its timing are needed to realize a given task.
    To demonstrate the effectiveness of our method, we applied it to the task of Wadaiko (traditional Japanese drum) drumming as an example, and verified the best configuration of DTXNET.
  }%
  {%
    柔軟な身体はその劣駆動性から環境接触等に対して優れている反面その制御は難しく, 与えられたタスクを動的かつ正確に実行するための手法を開発していく必要がある.
    我々は柔軟な身体における制御指令列による身体変化と, それにより実行されるタスクの変化を汎用的に記述する再帰的ネットワークを構築し, それを用いることで最適な制御指令を計算する.
    本研究は様々なマニピュレータ・タスクに適用可能であるが, 一つの例として太鼓を叩くという疎なイベント情報のみを出力する, 既存手法では扱うことが難しいタスクを設定し, 本ネットワークにより意図したタイミングと音量を実現することで, その有効性を確認する.
  }%
\end{abstract}

\section{INTRODUCTION}\label{sec:introduction}
\begin{figure}[t]
  \centering
  \includegraphics[width=0.9\columnwidth]{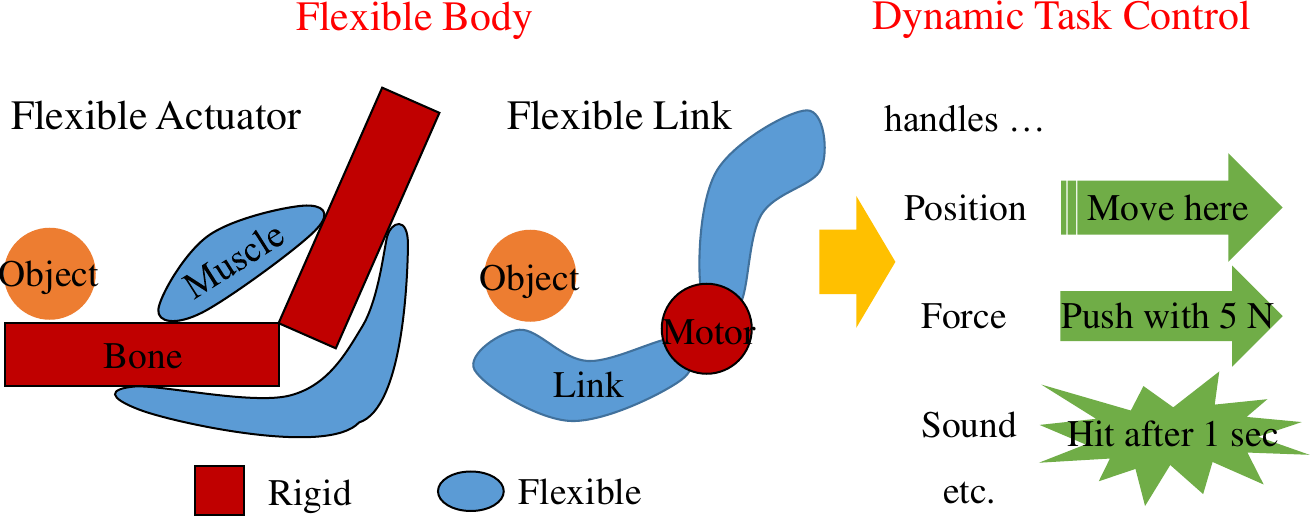}
  \caption{Our goal: dynamic task control of the flexible body.}
  \label{figure:motivation}
  \vspace{-3.0ex}
\end{figure}

\switchlanguage%
{%
  Conventionally, the increase of rigidity has been emphasized to achieve precise control of the robot body \cite{hirai1998asimo}.
  On the other hand, in recent years, the necessity and effectiveness of the soft robot, whose body is underactuated and flexible, are being reconfirmed \cite{kim2013softrobotics, lee2017softrobotics}.
  It has advantages in environmental contact, including human-robot interaction and the impact of falling down.

  The flexible body is difficult to control, especially when applying widely-used conventional control methods, e.g. quadratic programming \cite{escande2014qp} and model predictive control \cite{wieber2006mpc}.
  It is because they have been designed for the rigid body; they mostly handle the states, which are calculated from robot postures, such as the center of gravity and contact state.
  As for the flexible body, the posture cannot be determined by just the output of actuators, and the modeling of its body is difficult.
  Furthermore, although the conventional methods can be applied for tasks such as applying force and moving objects to a certain degree, they are not applicable for tasks of handling sparse event information such as drumming, hitting with a hammer, and playing tennis.
  This is because ordinary motion equations are not adequate to represent sparse task state, and therefore, it is difficult to give successive feedback signals for the sparse task.

  There are many studies regarding the control of the flexible body \cite{kiang2015flexible}.
  Methods to control a robot with flexible links by precise modeling have been developed since over 30 years ago \cite{book1984flexible, kotnik1988acceleration}.
  However, because the modeling of the flexible body is difficult, methods that do not require precise modeling have been developed, e.g. using fuzzy control \cite{moudgal1995flexible}, reinforcement learning \cite{pradhan2012flexible}, and neural network \cite{su2001flexible}.
  The focuses of these methods are on the control of the arm tip and inhibition of vibration, and not on realizing given tasks.
  Also, they need to modelize the flexible body to a certain degree and search its parameters, so it is difficult to handle materials with unknown characteristics or octopus-like flexible robots.

  Controlling the flexible body has some common points with manipulating flexible objects, and so there are many related works.
  Inaba, et al. manipulated a rope with vision feedback \cite{inaba1987rope}.
  Yamakawa, et al. succeeded in dynamically manipulating flexible objects with precise modeling by high-speed robot hands \cite{yamakawa2011folding}.
  Also, in recent years, thanks to the remarkable growth of deep learning, several methods acquiring manipulation of flexible objects by trial and error such as EMD Net \cite{tanaka2018emd} and Dynamics-Net \cite{kawaharazuka2019dynamic} have been proposed.
  There are also methods to learn the manipulation of flexible objects \cite{yang2017repeatable} and even how much force to apply \cite{lee2015learning} from teaching signals.
  These methods represent the state of flexible objects, which is difficult to modelize, by image or point cloud, and make use of them for the control of the actual robot.
  While the manipulation of flexible objects aims to change their posture and shape, the flexible body control aims to not only change its posture but also realize given tasks.
  There is also a method to learn task execution using image by a robot with flexible joints \cite{wu2018flexible}, but it is only verified at the simulation level and needs teaching signals.
}%
{%
  柔軟な身体を制御することは難しく, これまでのロボティクスは剛性を高め, 精度を高くすることで目的のタスクを実現するものがほとんどだった\cite{hirai1998asimo}.
  近年, 環境接触や転倒による衝撃, 人間との共生等から, これまでの剛なロボティクスが見直され始めており, その劣駆動で本質的に柔軟な身体を模索するソフトロボティクスの重要性が再確認され始めている\cite{kim2013softrobotics, lee2017softrobotics}.
  しかし, 柔軟な身体の制御には, 従来の剛なロボットにおける制御手法を適用することが難しい.
  例えば, 剛なロボットにおいては, 与えられたタスクを実行するためのロボットの姿勢を, 2次計画法\cite{escande2014qp}やモデル予測制御\cite{wieber2006mpc}等を用いて計算する手法が多く開発されてきた.
  これらはロボットの姿勢等から直接求解可能なタスク状態のみを扱い, 重心位置や接触状態等を操作するものが多い.
  しかし, 柔軟な身体は劣駆動性からアクチュエータの出力により姿勢が一意に定まらず, 姿勢等からタスクの状態を決定することも難しいため, これらの適用は困難である.
  また, これらの手法は力を加えたり, 物体の位置を操作したりするタスクを扱うことはできるかもしれないが, 音のような時間的に疎な情報を扱うタスクを行うことはより難しい.
  これは, 通常の状態方程式ではこれらを表現することが難しく, また, 報酬が疎なためそのタスクを達成するまでの連続的なフィードバックを与えることが難しいことが原因である.
  よって, 柔軟な身体を動的かつ正確に制御し, 目的のタスク状態を直接指定するのみで, 力や位置だけでなく, 音のような疎な情報も統一的に実現することのできる手法を開発する必要がある(\figref{figure:motivation}).

  これまで, 多くの柔軟な身体を制御するための手法が開発されてきた\cite{kiang2015flexible}.
  古くから, 柔軟なリンクを持つアームを, 完全なモデリングにより制御する手法が開発されている\cite{book1984flexible, kotnik1988acceleration}.
  しかし, 柔軟な身体の困難なモデリングを解決するため, Fuzzy制御を用いた手法\cite{moudgal1995flexible}, 強化学習を用いた手法\cite{pradhan2012flexible}, ニューラルネットワークを用いた手法\cite{su2001flexible, sum2017flexible}等の制御も開発され続けている.
  これらの手法の多くは, アーム先端の位置を制御すること, 柔軟な身体の振動を抑制することに注力しており, 柔軟な身体の姿勢だけでなく, 定められたタスクを動的に実現するものではない.
  そして, 柔軟な身体はある程度モデル化され, それらのパラメータを学習的に探索するものがほとんどであり, 身体がタコのような柔軟な身体であったり, よりモデル化困難な素材を使った場合には適用が難しい.

  また, 柔軟な身体を制御することは, 柔軟物体を操作することと多くの共通した問題を扱っている.
  稲葉らは視覚フィードバックによりロープを操り\cite{inaba1987rope}, 山川らは柔軟物体をモデリングして高速ロボットハンドを用いて動的に操作することに成功している\cite{yamakawa2010knotting, yamakawa2011folding}.
  また近年, 深層学習の発展は著しく, EMD Net \cite{tanaka2018emd}やDynamics-Net \cite{kawaharazuka2019dynamic}等, 試行錯誤により柔軟物体操作を獲得する手法が提案されている.
  教示により柔軟物体操作を学習する例も存在し\cite{yang2017repeatable}, 柔軟物体の姿勢操作だけでなく力のかけ方を学習する手法も存在する\cite{lee2015learning}.
  これらは, モデリング困難な柔軟物体を画像やPoint Cloudを用いることで記述し, 実ロボットの制御に役立てている.
  柔軟物体操作は操作物体の姿勢や状態を遷移させていくのに対して, 柔軟身体操作は自身の姿勢だけでなく, 最終的な目標は与えられたタスクを実現することである.
  画像を用いて柔軟関節を持つロボットのタスク実行を学習させる手法\cite{wu2018flexible}も存在するが, シミュレーションのみの実行であり, 教示信号を要する.
}%

\begin{figure*}[t]
  \centering
  \includegraphics[width=1.9\columnwidth]{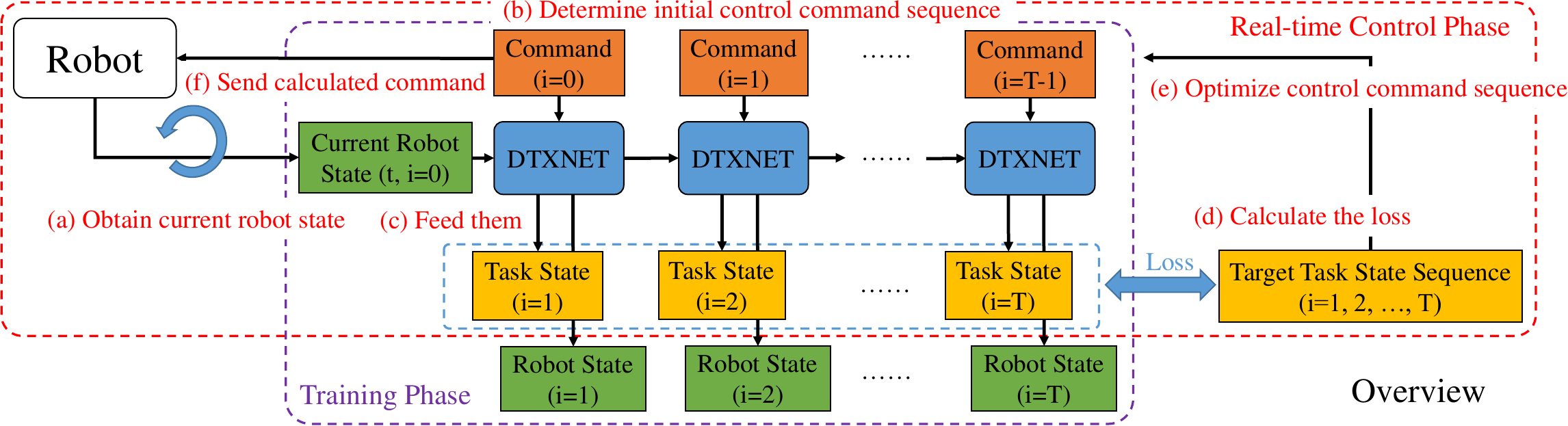}
  \caption{Basic structure of DTXNET and its application to a real-time control system.}
  \label{figure:network-structure}
  \vspace{-3.0ex}
\end{figure*}

\switchlanguage%
{%
  In this study, we propose a method controlling the flexible body precisely and dynamically to realize given tasks, which handle not only force and position but also sparse event information (\figref{figure:motivation}).
  Our method does not require any teaching or demonstration, and works by merely commanding the target task state.
  Our method consists of Dynamic Task Execution Network (DTXNET) and real-time backpropagation-based calculation of control command using DTXNET.
  DTXNET is a deep neural network which represents the motion equation of the flexible body and task using not only its joint angle, velocity, and torque but also image information.
  In order to make use of the dynamic characteristics of the flexible body, joint torque, which can directly control acceleration, is used as the control command.
  Our method is applicable to various manipulators and tasks if sensor information of actuators and image are available and the task state is clearly defined, because DTXNET does not require any task-specific structure.
  As one example, we realize a difficult task handling sound, which temporally outputs sparse events, instead of a task applying force or moving objects which can be solved to a certain degree by conventional frameworks.

  The detailed contributions of this study are,
  \begin{itemize}
    \item Structure of DTXNET using image information to realize dynamic tasks by a flexible manipulator.
    \item Real-time control method to calculate optimized torque command using DTXNET and its backpropagation through time.
    \item Realization of a dynamic task of handling sparse event information.
  \end{itemize}

  In the following sections, first, we will explain the basic structure of DTXNET, various configurations of DTXNET by changing inputs and outputs, training phase of DTXNET, and real-time control phase using DTXNET.
  Next, we will set a task of Wadaiko (traditional Japanese drum) drumming using a flexible manipulator, and compare the performance among various configurations of DTXNET by changing the input and output of the robot state.
  Finally, we will discuss the network structure and experiments, and state the conclusion.
}%
{%
  そこで本研究では, \figref{figure:motivation}に示すコンセプトのように, モデリング困難な柔軟なアームを用いて, 教示を用いることなく動的に所望のタスクを実現する手法を開発する.
  柔軟な身体とタスクの運動方程式を関節の角度・速度・トルク情報だけでなく, 画像情報を用いて記述するDynamic Task Execution Network (DTXNET)を開発する.
  柔軟な身体の動的特性を活かすために, 関節角度や関節速度ではなく, 加速度を直接操作可能な関節トルクによる制御を行う.
  本手法はアクチュエータまたは画像情報が取得可能であり, 扱いたいタスクの状態を定義することができれば, どのようなマニピュレータに対しても汎用的に適用可能な手法である.
  その一例として, 物体に力を加えるなど連続値を扱う, これまでの状態方程式を用いた枠組みである程度解決可能なタスクではなく, 音という時間的に疎なイベント情報を扱うより難しいタスクを実現することを試みる.
  本研究の詳細なコントリビューションを以下に示す.
  \begin{itemize}
    \item 柔軟な身体により動的にタスクを実現するための画像入力を用いたDTXNETの構成法
    \item DTXNETを展開し通時的誤差逆伝播法\cite{rumelhart1986bptt}を用いることで最適制御入力を算出するリアルタイム制御法
    \item 音という時間的に疎なイベント情報を扱う動的タスクの実現
  \end{itemize}

  以降ではまず, DTXNETの構成と, 入力と出力によるDTXNETの構造分類, DTXNETの訓練方法とそれを用いたリアルタイム制御手法について述べる.
  次に, 柔軟マニピュレータを用いて太鼓を叩くという動作を設定し, DTXNETの状態入力や状態出力を変化させた様々な構造において, タスク達成度を比較する.
  最後に, これらネットワークと実験について考察し, 結論を述べる.
}%

\section{DTXNET and Real-time Control System} \label{sec:dte-net}
\switchlanguage%
{%
  In \secref{subsec:basic-structure} - \secref{subsec:optimized-torque}, we will explain the overview of DTXNET and a real-time control system using it, and in \secref{subsec:whole-system}, we will explain the specific implementation and parameters for experiments.
}%
{%
  本節では初めは一般化してネットワーク等を記述し, 最後に具体的な実装やパラメータについて述べる.
}%
\subsection{Basic Structure of DTXNET} \label{subsec:basic-structure}
\switchlanguage%
{%
  We show the overview of DTXNET and its application to the dynamic control system in \figref{figure:network-structure}.
  DTXNET is a versatile network which represents the transitions of the current robot state and task state due to control commands.
  This network does not depend on the kind of manipulators and tasks.
  The equations of the network are,
  \begin{align}
    \bm{h}(i=0) &= \bm{f}_{init}(\bm{s}_{init}(t)) \label{eq:init} \\
    \bm{h}(i+1) &= \bm{f}_{update}(
    \begin{bmatrix}
      \bm{h}^{T}(i) & \bm{u}^{T}(i)
    \end{bmatrix}^{T}
    ) \label{eq:update} \\
    \bm{s}(i+1) &= \bm{f}_{robot}(\bm{h}(i+1)) \label{eq:state} \\
    \bm{o}(i+1) &= \bm{f}_{task}(\bm{h}(i+1)) \label{eq:out}
  \end{align}
  where $\bm{s}$ is the measured robot state, $\bm{o}$ is the task state, $\bm{u}$ is the control command, and $\bm{h}$ is the hidden state that is obtained by embedding the robot state.
  $t$ is the time step for the robot, $i$ is the time step for DTXNET, and the interval of these two variables is the same.
  $\bm{f}_{init}(t)$ is a function embedding the measured robot state to the hidden state, $\bm{f}_{update}$ is a function that outputs the hidden state at a next time step from the current hidden state and control command, and $\bm{f}_{robot}$ and $\bm{f}_{task}$ are functions that output the robot state and task state from the hidden state, respectively.
  $\bm{s}_{init}(t)$ and $\bm{s}(i+1)$ can represent different contents; for example, the former can be position and velocity and the latter can be just position.
  $\bm{s}(i+1)$ should be derived from $\bm{s}_{init}(t)$.

  We use actuator and image information as the robot state, and joint torque as the control command.
  If we assume the robot body is rigid, we can express $\bm{s}$ only by its joint angle and velocity.
  On the other hand, for the flexible body, we need to include image information and joint torque in $\bm{s}$.
  When considering the flexible body as an underactuated multi-link structure, its state can be expressed by image information.
  Also, although the control command $\bm{u}$ is joint position or velocity for an ordinary position control robot, in this study, we use joint torque command, which can directly control the acceleration, in order to achieve a truly dynamic control.
}%
{%
  本研究で提案するDTXNETの構造とそれを用いた動的制御の概要を\figref{figure:network-structure}に示す.
  DTXNETは, ある初期状態において制御入力を加えていくと, ロボットの身体状態, そしてタスクの状態が変化していくことを表す汎用的なネットワークである.
  本手法は, ロボットの形やタスクの種類等に依存せず, 汎用的にロボット状態, 制御入力, タスク状態の関係を記述していくものとなる.
  これは式で表すと以下のようになる.
  \begin{align}
    \bm{h}(i=0) &= \bm{f}_{init}(\bm{s}_{init}(t)) \label{eq:init} \\
    \bm{h}(i+1) &= \bm{f}_{update}(
    \begin{bmatrix}
      \bm{h}^{T}(i) & \bm{u}^{T}(i)
    \end{bmatrix}^{T}
    ) \label{eq:update} \\
    \bm{s}(i+1) &= \bm{f}_{robot}(\bm{h}(i+1)) \label{eq:state} \\
    \bm{o}(i+1) &= \bm{f}_{task}(\bm{h}(i+1)) \label{eq:out}
  \end{align}
  ここで, $\bm{s}$はロボットの観測される身体状態, $\bm{o}$はあるタスクの状態, $\bm{u}$は制御入力, $\bm{h}$はロボットの圧縮された潜在的な状態を表す.
  $t$と$i$は異なり, 前者はロボットにおける時間ステップ, 後者はDTXNETにおける時間ステップを指し, 両者のステップ間隔は同じである.
  $\bm{f}_{init}$はロボットの観測される身体状態を潜在状態に落としこむ関数, $\bm{f}_{update}$は潜在状態と制御入力から次のステップにおける潜在状態を出力する関数, $\bm{f}_{robot}$, $\bm{f}_{task}$は潜在状態からそれぞれ次ステップで観測されるべきロボットの身体状態, タスク状態を出力する関数である.
  $\bm{s}_{init}(t)$と$\bm{s}(i+1)$は同じ形である必要はなく, 例えば前者は位置と速度, 後者は位置のみ, 等のバリエーションがあり得るが, $\bm{s}_{init}(t)$の方が$\bm{s}(i+1)$に比べて情報量が多くなければならない.

  本研究では, このネットワークを用いて柔軟な身体を動的に制御することを考える.
  通常の剛体を仮定したロボットであれば, $\bm{s}$は関節角度・関節角速度のみで表現できるが, 柔軟な身体を制御するためには, それらだけでは表現できない身体状態が存在する.
  そこで本研究では, 関節角度・関節角速度に加え, 関節トルク・画像を$\bm{s}$として加える.
  柔らかい身体を無限の自由度を持ったリンクと考えれば, 画像によってその身体状態を表現可能なことがわかる.
  しかし, それでも$\bm{s}$には含まれないロボットにおける隠れ状態が存在する可能性があると考え, DTXNETにおける$\bm{f}_{update}$は長期的な状態の記憶が可能なLong Short-Term Memory (LSTM) \cite{hochreiter1997lstm}を用いて表す.
  また, 通常の位置制御ロボットであれば制御入力$\bm{u}$は関節角度または関節速度であるが, 真に動的な柔軟身体の制御を目指し, 本研究では直接加速度項を制御可能な関節トルクを制御入力として用いる.
}%

\begin{figure*}[t]
  \centering
  \includegraphics[width=1.8\columnwidth]{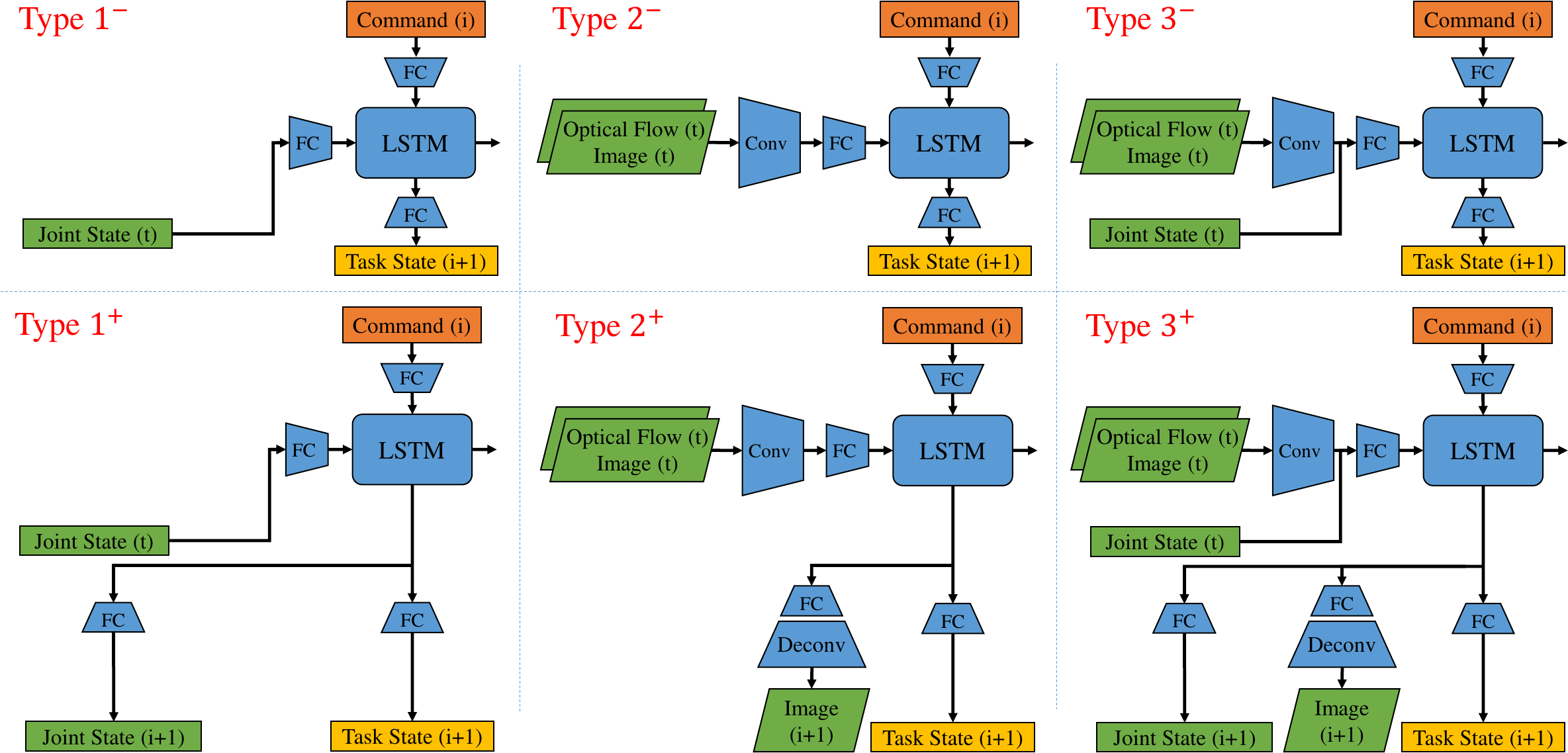}
  \caption{Various configurations of DTXNET by the design of the robot state: using only Joint State (Type 1), using only Image (Type 2), or using both of them (Type 3), and by the output of the network; Type $\cdot^-$ does not output the robot state, and Type $\cdot^+$ outputs the robot state.}
  \label{figure:various-network}
  \vspace{-3.0ex}
\end{figure*}

\subsection{Various Configurations of DTXNET} \label{subsec:various-network}
\switchlanguage%
{%
  We represent \equref{eq:init} - \equref{eq:out} as neural networks.
  We can consider six types of configurations of DTXNET as shown in \figref{figure:various-network}.
  DTXNET must fulfill the minimum requirements of inputting the measured robot state and control command to DTXNET and outputting the task state.
  In these configurations, Joint State represents a concatenated vector of the information of joint angle, velocity, and torque.
  FC, Conv, Deconv, and LSTM represent fully connected layers, convolutional layers, deconvolutional layers, and Long Short-Term Memory \cite{hochreiter1997lstm}, respectively.
  Although there are several choices for the recurrent network structure, in this study, we used LSTM, which is one of the most representative recurrent networks.

  There are three options in the design of the robot state: using only Joint State (Type 1), using only Image (Type 2), or using both Joint State and Image (Type 3).
  In these configurations, Image means the current image and optical flow regarding the inputted $\bm{s}_{init}$, and only the current image regarding the outputted $\bm{s}$.
  Because we knew that the inference of the optical flow is difficult based on our preliminary experiments, there is no output of the optical flow.

  There are two options in the output of the network.
  Type $\cdot^-$ does not output the robot state, and Type $\cdot^+$ outputs the robot state.
  While the robot state is not directly involved in the realization of the task, we can expect DTXNET to increase the prediction accuracy by using the robot state at training phase.

  According to the above classifications, the six types of DTXNET are obtained by the combination of these options: $1^-$, $2^-$, $3^-$, $1^+$, $2^+$, and $3^+$.

  In this study, we not only realize a dynamic task using DTXNET, but also consider how the difference in types of DTXNET influences the task achievement.
}%
{%
  \equref{eq:init} - \equref{eq:out}を単純にニューラルネットワークに置き換えることを考える.
  観測されうるロボットの身体状態と制御入力をネットワークの入力として加える, 与えられたタスクの状態を出力として加える, という最低限の制約を満たした\figref{figure:various-network}のような6種類のtypeを考えることができる.
  ここでは, 関節角度・関節角速度・関節トルクの情報をJoint Stateと表現する.
  FCは全結合層, Convは畳み込み層, Deconvは逆畳み込みを表す.

  まず, ロボットの観測される身体状態の種類によって分類する.
  type $1^-, 1^+$は身体状態がJoint Stateのみ, type $2^-, 2^+$は身体状態がImageのみ, type $3^-, 3^+$は身体状態がJoint StateとImageの場合である.
  ここでImageとは, 入力となる$\bm{s}_{init}$においては画像とオプティカルフローを指し, 出力となる$\bm{s}$においては画像のみを指す.
  実験からオプティカルフローを推論することは難しいと判断し, 出力にはオプティカルフローは含めていない.

  次に, ネットワークの出力として身体状態を出力するかどうかという分類を考えることができる.
  タスクを実行することが目的のため, 無用に身体状態を出力することはない.
  type $1^-, 2^-, 3^-$は身体状態を出力せず, type $1^+, 2^+, 3^+$は身体状態を出力する.

  本研究では, DTXNETを用いた動的タスク実現の手法だけでなく, これらのDTXNETの構造の違いがどのようにタスク実現に寄与するかについての考察も行う.
}%

\subsection{Training Phase of DTXNET} \label{subsec:training}
\switchlanguage%
{%
  Because both robot state and task state, which are the input and output of DTXNET, can be easily obtained, the training procedure of DTXNET does not require any manual annotations.
  First, we send random control commands to the robot, and store the measured robot state $\bm{s}$, control command $\bm{u}$, and task state $\bm{o}$.
  Then, we determine the number of time steps to expand DTXNET ($T_{train}$).
  After inputting $\bm{s}_{init}(t)$ into DTXNET, $\{\bm{s}(i=1), \cdots, \bm{s}(i=T_{train})\}$ and $\{\bm{o}(i=1), \cdots, \bm{o}(i=T_{train})\}$ are predicted by inputting $\bm{u}(i=c)\;(0 \leq c < T_{train})$ at $T_{train}$ times, and DTXNET is trained by calculating the loss ($L_{train}$) between the predicted and actual values.
  In this training phase, $T_{train}$ should be larger than the number of time steps to expand DTXNET by ($T_{control}$) at the real-time control phase.
}%
{%
  DTXNETは, 制御入力によるロボットの身体状態の遷移とタスク状態の遷移を記述するため, アノテーション等は必要としない.
  まず, ランダムな制御入力によりロボットを動作させ,　その際に観測されたロボットの身体状態$\bm{s}$, 制御入力$\bm{u}$, タスクの状態$\bm{o}$を蓄積しておき, これらを元に学習させる.
  DTXNETを展開する数$T_{train}$を決め, $\bm{s}_{init}(t)$をDTXNETに入力した後, $T_{train}$回$\bm{u}(i=c)\;(0 \leq c < T_{train})$を入力し, $\{\bm{s}(i=1), \cdots, \bm{s}(i=T)\}$, $\{\bm{o}(i=1), \cdots, \bm{o}(i=T)\}$を予測し, 実際の値とloss ($L_{train}$)を取って学習させる.
  ここで$T_{train}$は後にリアルタイム制御でDTXNETを展開するステップ数$T_{control}$より大きいことが望ましい.
  本研究では更新則としてAdam \cite{kingma2015adam}を用いている.
}%

\subsection{Real-time Control Phase Using DTXNET} \label{subsec:optimized-torque}
\switchlanguage%
{%
  The real-time control phase using trained DTXNET has six steps, as shown in \figref{figure:network-structure}.
  \begin{enumerate}
    \item Obtain the current robot state from the actual robot
    \item Determine the initial control command sequence
    \item Feed them into DTXNET
    \item Calculate the loss
    \item Optimize the control command sequence
    \item Send the calculated control command to the robot
  \end{enumerate}
  All steps from (a) to (f) are executed within one time step that controls the actual robot.

  In (a), the current sensor information of the actual robot $\bm{s}_{init}(t)$ is obtained, and the current hidden state $\bm{h}(i=0)$ is calculated by \equref{eq:init}.

  In (b), the initial control command sequence is calculated before optimization.
  This procedure is important, because the final calculation result largely depends on the initial value.
  We define the control command sequence that is optimized at the previous time step, as $\bm{u}^{seq}_{pre} = \{\bm{u}_{pre}(i=-1), \bm{u}_{pre}(i=0), \cdots, \bm{u}_{pre}(i=T_{control}-3), \bm{u}_{pre}(i=T_{control}-2)\}$.
  By shifting it and replicating its last term $\bm{u}_{pre}(i=T_{control}-2)$, the initial control command sequence is constructed as $\bm{u}^{seq}_{init, 0} = \{\bm{u}_{pre}(i=0), \bm{u}_{pre}(i=1), \cdots, \bm{u}_{pre}(i=T_{control}-2), \bm{u}_{pre}(i=T_{control}-2)\}$.
  By starting from the previous optimized value, better control command can be efficiently obtained.
  Also, we define the number of data per minibatch as $N_{1}$, and the minimum and maximum value of control command as $\bm{u}_{min}$ and $\bm{u}_{max}$, respectively.
  We divide $[\bm{u}_{min}, \bm{u}_{max}]$ equally into $N_{1}-1$ parts and construct $N_{1}-1$ control command sequences $\bm{u}^{seq}_{init, k} (1 \leq k < N_{1})$ filled by each value.
  Then, a minibatch of $N_{1}$ initial control command sequences $\bm{u}^{seq}_{init, k} (0 \leq k < N_{1})$ is constructed.
  % In the following steps, we use the command sequence whose loss is the lowest in this minibatch.

  In (c), as stated in \secref{subsec:training}, we determine $T_{control}$, which is the number of time steps to predict, and update DTXNET by \equref{eq:update} $T_{control}$ times.
  While updating \equref{eq:update}, the task state is predicted by \equref{eq:out} regarding each $\bm{h}$.
  In this real-time control phase, we can omit \equref{eq:state} to reduce the computational cost, because we do not have a certain desired robot state and the prediction of the robot state cannot be used for the calculation of loss.
  This omission is important for the real-time computation.
  When \equref{eq:state} is omitted, the difference among the six types is only the calculation of $\bm{f}_{init}$.
  Since \equref{eq:init} is calculated only once at the beginning of $T_{control}$ times iteration, the computational cost is almost the same among the six types.

  In (d), the loss $L_{control}$ between the target task state sequence and predicted task sequence $\{\bm{o}_{i=1}, \cdots, \bm{o}_{i=T_{control}}\}$ is calculated.
  The implementation of $L_{control}$ depends on the task, which we will explain in \secref{subsubsec:loss-calculation}.

  In (e), first, $\bm{u}^{seq}=\{\bm{u}(i=0), \cdots, \bm{u}(i=T_{control}-1)\}$ with the lowest $L_{control}$ is determined in a minibatch constructed at step (b).
  $\bm{u}^{seq}$ is optimized from $L_{control}$ by using backpropagation through time \cite{rumelhart1986bptt}, as shown below,
  \begin{align}
    \bm{g} &= dL_{control}/d\bm{u}^{seq} \\
    \bm{u}^{seq}_{optimized} &= \bm{u}^{seq} - \gamma\bm{g}/|\bm{g}| \label{eq:opt}
  \end{align}
  where $\gamma$ is the learning rate of optimization, $\bm{g}$ is the gradient of $L_{control}$ regarding $\bm{u}^{seq}$, $\bm{u}^{seq}_{optimized}$ is $\bm{u}^{seq}$ after the optimization.
  We can determine the $\gamma$ manually, but in this study, a minibatch is constructed by changing $\gamma$ variously, and the best $\gamma$ is chosen.
  We define the number of data per minibatch as $N_{2}$ and the maximum value of $\gamma$ as $\gamma_{max}$.
  We divide $[0, \gamma_{max}]$ equally into $N_{2}$ parts, and a minibatch with $N_{2}$ control command sequences, which is $\bm{u}^{seq}_{optimized}$ optimized with each $\gamma$, is constructed.
  Steps (c) and (d) are executed again, and finally $\bm{u}^{seq}_{optimized}$ with the lowest loss $L_{control}$ is determined.

  In (f), $\bm{u}^{seq}_{optimized}$ is sent to the actual robot.
  Because the maximum calculation time to be in time for the control frequency is used, we send not $\bm{u}^{seq}_{optimized}(i=0)$ but $\bm{u}^{seq}_{optimized}(i=1)$.
}%
{%
  Training phaseで学習されたDTXNETを用いたリアルタイム制御は, \figref{figure:network-structure}に示すように6つの手順を踏む.
  \begin{enumerate}
    \item Obtain the current robot state from the actual robot
    \item Determine the initial control command sequence
    \item Feed them into DTXNET
    \item Calculate the loss
    \item Optimize control command sequence
    \item Send the calculated control command to the actual robot
  \end{enumerate}
  この全手順をロボットの制御周期ごとに行う.

  (a)では, 現在のロボットのセンサ情報$\bm{s}_{init}(t)$を取得し, \equref{eq:init}により現在の潜在状態$\bm{h}(i=0)$を算出する.

  (b)では, 最適化する前の制御入力の初期値を求める.
  このステップは重要であり, 初期解によって最終的な計算結果が大きく異なる.
  前ステップで最適化され最終的に実機に送られた制御入力列を$\{\bm{u}_{pre}(i=-1), \bm{u}_{pre}(i=0), \cdots, \bm{u}_{pre}(i=T_{control}-3), \bm{u}_{pre}(i=T_{control}-2)\}$とする.
  これを1ステップずつずらして$\bm{u}_{pre}(i=T_{control}-2)$を複製した, $\{\bm{u}_{pre}(i=0), \bm{u}_{pre}(i=1), \cdots, \bm{u}_{pre}(i=T_{control}-2), \bm{u}_{pre}(i=T_{control}-2)\}$を初期値に用いる.
  これは, 前回最適化された値を引き継ぐことで, より効率よくそのタスクを実現可能な制御入力を得るためである.
  また, 制御入力の最小値と最大値を$\bm{u}_{min}, \bm{u}_{max}$とする.
  この最小値と最大値の間を$N_{batch1}-1$個に分割し, $T_{control}$ステップ分全てその同じ値で埋めた制御入力列も作成し, 合わせて$N_{batch1}$個のバッチを作成する.
  これは, タスクの指令状態が変化したり, 状況が変化したりしたときに柔軟に対応できるようにするためである.
  以降では, これらのバッチの中から最もlossが下がった制御入力列を採用していくことになる.

  (c)では, \secref{subsec:training}でも述べた何ステップ先まで予見するかの値$T_{control}$を決定し, 制御入力を加えてDTXNETにおける\equref{eq:update}を更新していく.
  $T_{control}$回\equref{eq:update}を更新すると同時に, それぞれの$\bm{h}$に対して\equref{eq:out}によりタスク状態を予測する.
  ここで重要なのは, Training phaseでは\equref{eq:state}のlossも返す必要があるが, Control phaseではタスクの実現のみを考えるため, \equref{eq:out}のみ走らせれば良い.
  ゆえにControl phaseにおいては, これまで説明した6種類のtypeは\equref{eq:init}のみが異なり, 計算にかかる時間はほとんど変わらない.

  (d)では, targetとなるタスク状態と予測されたタスク状態$\{\bm{o}_{i=1}, \cdots, \bm{o}_{i=T_{control}}\}$のloss ($L_{control}$)を計算する.
  $L_{control}$の実装はタスクによって異なり, 本研究における具体的なlossの計算は後に述べる.

  (e)ではまず, (d)で計算された$N_{batch1}$個のそれぞれのバッチの損失の中でも最も$L_{control}$が小さな$\bm{u}^{seq}=\{\bm{u}_{i=0}, \cdots, \bm{u}_{i=T_{control}-1}\}$を決定する.
  その$L_{control}$を元に以下のように通時的誤差逆伝播法\cite{rumelhart1986bptt}を用いて$\bm{u}^{seq}$を更新していく.
  \begin{align}
    \bm{g} &= dL_{control}/d\bm{u}^{seq} \\
    \bm{u}^{seq}_{optimized} &= \bm{u}^{seq} - \gamma\bm{g}/|\bm{g}| \label{eq:opt}
  \end{align}
  ここで, $\gamma$は更新に関する係数, $\bm{g}$は$L_{control}$の$\bm{u}^{seq}$に対する勾配, $\bm{u}^{seq}_{optimized}$は更新後の$\bm{u}^{seq}$である.
  このとき, $\gamma$の値を決め打ちしても良いが, 本研究では様々な$\gamma$によってバッチを作成し, 最も良い$\gamma$を選ぶ.
  $\gamma$の最大値$\gamma_{max}$を決め, 0から$\gamma_{max}$までの値を$N_{batch2}$等分し, それらによって更新された$\bm{u}^{seq}_{optimized}$を$N_{batch2}$個作成する.
  もう一度(c)と(d)を行い, 最も最終的な$L_{control}$が小さかった$\bm{u}^{seq}_{optimized}$を決定する.

  (f)では, $\bm{u}^{seq}_{optimized}$を実機に送る.
  (a)から(e)は制御周期に間に合う最大限の時間を使うため, ここで実機に送るのは$\bm{u}^{seq}_{optimized}(i=0)$ではなく, $\bm{u}^{seq}_{optimized}(i=1)$となる.
}%

%\begin{figure}[t]
%  \centering
%  \includegraphics[width=1.0\columnwidth]{figs/optimization}
%  \caption{Details of the generation of initial control command sequences at step (b) and their optimization at step (e).}
%  \label{figure:optimization}
%  \vspace{-3.0ex}
%\end{figure}

\subsection{Implementation Details} \label{subsec:whole-system}
\switchlanguage%
{%
  We will explain the detailed implementation of our system.
  We will explain the sound processing of Wadaiko, image processing, detailed network structure, parameters for the training and control phase, and calculation of loss, in order.
}%
{%
  詳細な実装について述べる.
  本研究で扱う太鼓を叩くタスクに必要な音声処理, 画像処理, ネットワークの具体的な構造, Training phaseやControl phaseにおけるパラメータ, lossの計算方法について順に述べていく.
}%

\subsubsection{Sound Processing} \label{subsubsec:sound-processing}
\switchlanguage%
{%
  In this study, we set the target state of Wadaiko as a 1-dimensional value of the sound volume.
  First, the sound spectrum in each frequency is calculated by fast Fourier transform.
  We determine the frequency band with the largest spectrum when drumming Wadaiko (in this study, 270 - 310 Hz), and the largest spectrum value $v$ in the band is calculated.
  We determine the threshold $v_{thre}$ to judge whether it is noise or not.
  Then, the value $v$ if $v \geq v_{thre}$ or 0 if $v < v_{thre}$ is used as a task state.
}%
{%
  本研究では, 太鼓を叩くタスクにおける状態を, 音の大きさの一次元としている.
  まず, 得られた音声を高速フーリエ変換し周波数領域ごとの音の大きさを求める.
  太鼓が叩かれた時に最も大きなスペクトルが現れる周波数領域を決定し(本研究では270-310Hzとしている), その周波数帯でのスペクトル$v$の大きさを取得する.
  その音が雑音かどうかを判定する値$v_{thre}$を決め, それ以上であればその値$v$, それ以下であれば-1をタスクの状態として用いる.
}%

\subsubsection{Image Processing} \label{subsubsec:image-processing}
\switchlanguage%
{%
  The image is used after resizing and binarization.
  The detailed procedures are, in order, resizing, background subtraction, blurring, binarization, closing, and opening.
  A 640 $\times$ 480 image is resized to 64 $\times$ 64.
}%
{%
  画像は二値化処理を行ってから用いる.
  その処理はサイズの縮小, 背景差分処理, ぼかし, 閾値を決めて二値化, 膨張収縮である.
  サイズは, 640x480の画像を64x64xの画像にまでリサイズしている.
}%

\subsubsection{Detailed Network Implementation} \label{subsubsec:network-implementation}
\switchlanguage%
{%
  We will explain the detailed network structure of Type $3^+$ as an example.
  Regarding other types, the network does not use the part without correlation.

  First, we will explain \equref{eq:init}.
  A 12288-dimensional Image of 3 $\times$ 64 $\times$ 64 is embedded to 256 dimensions by convolutional layers.
  The convolutional layers have six layers, and the numbers of channels are 3 (input), 4, 8, 16, 32, and 64, respectively.
  Regarding all layers, the kernel size is 2 $\times$ 2, stride is 2 $\times$ 2, and padding is 1.
  We insert Batch Normalization \cite{ioffe2015batchnorm} after all the convolutional layers.
  The 3 channels of the input are for the current image and $xy$ directions of the optical flow.
  The Joint State and 256 dimensions of Image are embedded to 128 dimensions of $\bm{h}(i=0)$ by one fully connected layer.

  Next, we will explain \equref{eq:update}.
  It is expressed by one layer of LSTM in which both the input and output have 128 dimensions.
  In the procedure (c) stated in \secref{subsec:optimized-torque}, the hidden state of LSTM is initialized to $\bm{h}(i=0)$ calculated by \equref{eq:init} and the cell state of LSTM is initialized to 0.
  The control command $\bm{u}(i)$ is arranged to 128 dimensions by one fully connected layer, and is inputted in LSTM.

  \equref{eq:out} is composed of one fully connected layer, and converts $\bm{h}(i+1)$ to the task state.

  Finally, we will explain \equref{eq:state}.
  Joint State is predicted from $\bm{h}(i+1)$ through one fully connected layer.
  The image is predicted by six deconvolutional layers which have the same structures as the convolutional layers of \equref{eq:init}.
  Compared to the convolutional layers, in the last layer, the deconvolutional layers do not include Batch Normalization and the activation function is not ReLU but Sigmoid.

  In this study, we implement DTXNET by Chainer \cite{tokui2015chainer}, and the entire system runs on just CPU.
}%
{%
  具体的なネットワークの構造について述べるが, 基本的にtype $3^{+}$について述べる.
  その他のtypeに関しては, 例えば画像入力がない場合は, それを入力とするネットワークが使用されないような構造を取る.

  まず\equref{eq:init}について述べる.
  画像は畳み込みによって64x64の4096次元から256次元まで圧縮される.
  画像の畳み込み層は6層構造となっており, チャネル数はそれぞれ3 (input), 4, 8, 16, 32, 64とし, 全層においてKernel Size: 3x3, Stride: 2x2, Padding: 1とし, 全畳み込みの後にBatch Normalization \cite{ioffe2015batchnorm}を行う.
  入力の3チャンネルは現在画像とオプティカルフローのxy方向である.
  Joint Stateと画像の256次元を全結合層により128まで圧縮し$\bm{h}(i=0)$とする.

  次に, \equref{eq:update}について述べる.
  これは, 入力を128次元, 出力を128次元とするような一層のLSTMを用いて表現する.
  最初のステップにおいては, LSTMの隠れ状態を\equref{eq:init}により出力された128次元の$\bm{h}(i=0)$とし, LSTMのセル状態を0で初期化する.
  制御入力$\bm{u}(i)$は一層の全結合層により128次元に整形され, LSTMに入力される.

  \equref{eq:out}に関して, これは一層の全結合層からなり, 128次元の$\bm{h}(i+1)$をタスク状態に変換する.

  最後に, \equref{eq:state}について述べる.
  Joint Stateは$\bm{h}(i+1)$を一層の全結合層を通した後得られる.
  Imageは, \equref{eq:init}における画像の畳込み層の逆の構造をしており, 6層の逆畳み込みにより画像を復元する.
  畳み込み層と違う点は, 最終層にBatch Normalizationを含まず, 最終層の活性化関数のみReLUではなくSigmoidであるという点のみである.

  本研究ではこのDTXNETをChainer \cite{tokui2015chainer}により実装し, 本研究の全システムはCPUのみにより行われている.
}%

\subsubsection{Parameters of Training and Control Phase} \label{subsubsec:parameters}
\switchlanguage%
{%
  For the task of Wadaiko drumming, we set the control frequency to 15 Hz.
  We set $T_{control} = T_{train} = 8$; it means that DTXNET can predict about 0.53 sec ahead.
  Also, we set $N_{batch1} = N_{batch2} = 5$, and $\gamma_{max}=0.3$.
  The parameters stated above depend on the limit of calculation time; they should be adjusted depending on the performance of machine learning library and the specification of PC.
  We set $u_{min} = -0.3$ [Nm] and $u_{max} = 0.3$ [Nm] for all motors.
  We use Adam \cite{kingma2015adam} as an optimizer of DTXNET.
}%
{%
  まず, 本研究における太鼓叩きタスクの制御周期は15 Hzで行う.
  $T_{control} = T_{train} = 8$とし, 約0.53秒先までを予測する.
  また, $N_{batch1} = N_{batch2} = 5$, $\gamma_{max}=0.3$とする.
  上述の制限は, 計算量から来るものがほとんどであり, 用いるライブラリやPCの性能によって調整が可能である.
  全モータのトルク下限と上限は揃え, $u_{min} = -0.3$ [Nm], $u_{max} = 0.3$ [Nm]とする.
}%

\subsubsection{Loss Calculation} \label{subsubsec:loss-calculation}
\switchlanguage%
{%
  The loss $L_{train}$ at training phase is calculated as below,
  \begin{align}
    L_{train} = L_{j} + L_{i} + L_{o}
  \end{align}
  where $L_{j}$ is the loss of Joint State, $L_{i}$ is the loss of Image, and $L_{o}$ is the loss of task state.
  $L_{j}$ and $L_{i}$ are mean squared error of all elements, and they are adjusted by gains of $w_{j}=1.0$ and $w_{i}=30.0$.

  Regarding $L_{o}$, we control timing and volume of the sound to follow the target value.
  However, the sound information is sparse, and if the network learns it simply by mean squared error, a state of no sound output may produce the least loss.
  To solve this problem, we use the loss shown as below,
  \begin{align}
    &if\;\; o_{target} > 0 \notag\\
      &\;\;\;\;\;\;\;\;L_{o} = \alpha(o_{target}-o_{predicted})^2 \notag\\
    &else \notag\\
      &\;\;\;\;\;\;\;\;L_{o} = (o_{target}-o_{predicted})^2 \label{eq:loss}
  \end{align}
  where $o_{predicted}$ is the predicted sound volume, $o_{target}$ is the target sound volume (at training phase, this is the actual data to be predicted), and $\alpha (\alpha > 1)$ is a gain.
  Because the sound information is sparse, the loss is multiplied by $\alpha$ (in this study, $\alpha=10$) when making a sound.
  Also, at training phase, in order to adjust the scales of $L_{i}$, $L_{j}$, and $L_{o}$, $L_o$ is multiplied by a gain of $w_{o}=10.0$.
  At control phase, $L_{control}$ equals to $L_{o}$.
}%
{%
  Training phaseにおけるloss ($L_{train}$)は以下のように計算される.
  \begin{align}
    L_{train} = L_{j} + L_{i} + L_{o}
  \end{align}
  ここで, $L_{j}$はJoint Stateに関するloss, $L_{i}$は画像に関するloss, $L_{o}$はタスクの状態に関するlossを指す.
  $L_{j}, L_{i}$は全要素に対して平均二乗誤差を取ったものであり, その後それぞれ$w_{j}=1.0, w_{i}=30.0$の係数をかけている.

  $L_{o}$に関して, 本研究では音のタイミング, 大きさを指令した値に近づけるような制御を行う.
  しかし, 音がなるタイミングは非常にスパースであり, そのまま学習させてしまうと, 一切音がならない方が良い, というような結果になることがある.
  そこで, 以下のようなlossを用いる.
  \begin{align}
    &if\;\; o_{target} > 0 \notag\\
      &\;\;\;\;\;\;\;\;L_{o} = \alpha(o_{target}-o_{predicted})^2 \notag\\
    &else \notag\\
      &\;\;\;\;\;\;\;\;L_{o} = (o_{target}-o_{predicted})^2 \label{eq:loss}
  \end{align}
  ここで, $o_{predicted}$は予測された音の値, $o_{target}$は正しい値(Training phaseにおいては予測されるべき値, Control phaseにおいては指令値を指す), $\alpha (\alpha > 1)$は係数を表す.
  音が出る瞬間はスパースなため, 音が出る際はlossを$\alpha$倍している(本研究では$\alpha=10$).
  また, Training Phaseの際は, $L_{i}, L_{j}$と大きさを揃えるため, 10倍したうえで用いている.

  Control phaseにおいては, $L_{control} = L_{o}$となる.
}%

\section{Experiments} \label{sec:experiments}

\begin{figure}[t]
  \centering
  \includegraphics[width=1.0\columnwidth]{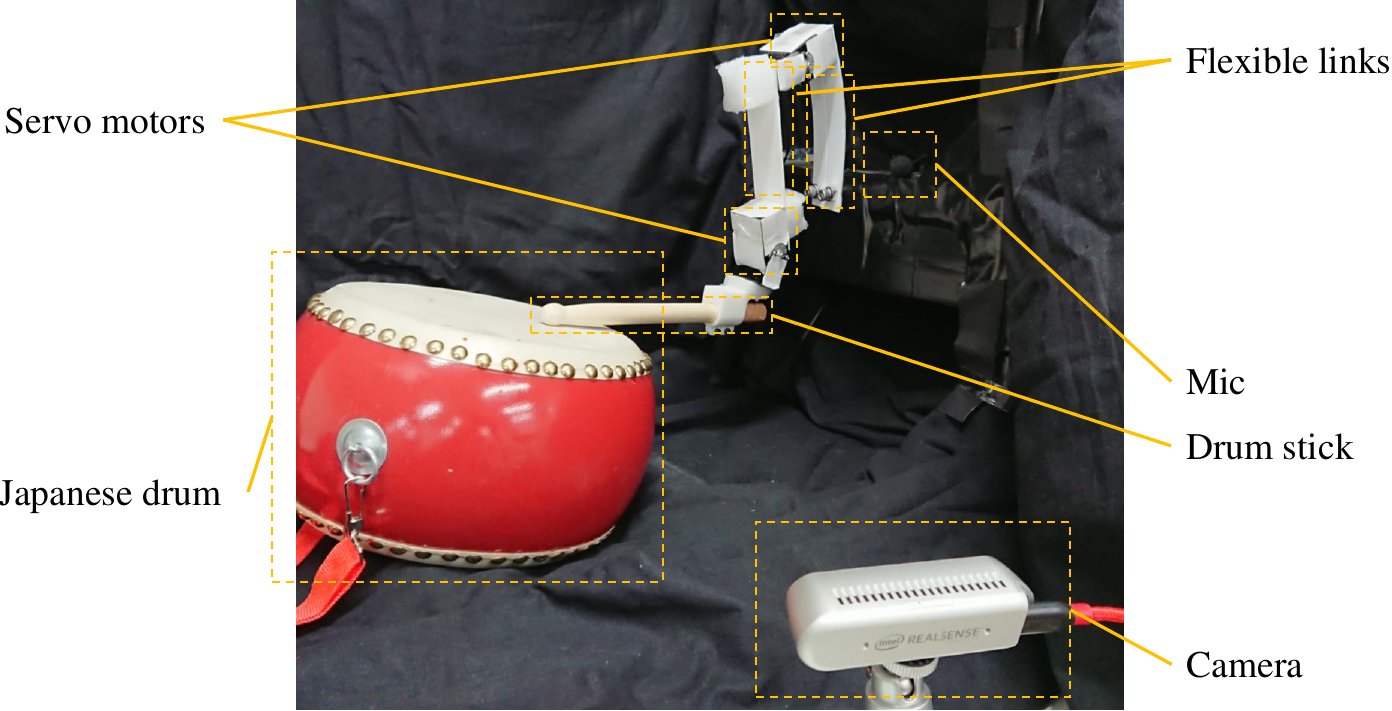}
  \caption{Experimental setup of Wadaiko drumming.}
  \label{figure:experimental-setup}
  \vspace{-3.0ex}
\end{figure}

\subsection{Experimental Setup} \label{subsec:experimental-setup}
\switchlanguage%
{%
  The experimental setup of Wadaiko drumming is shown in \figref{figure:experimental-setup}.
  The active degrees of freedom (DOFs) of the flexible manipulator is 2, and they are actuated by Dynamixel Motors (XM430-W210-R).
  We use D435 (Intel Realsense) as RGB Sensor, and set a mic next to the manipulator.
  The links of the flexible manipulator are composed of a soft cutting board.
  While the flexible body structure can absorb the impact of drumming, its control is difficult because the volume and timing of sound are different depending on the degree it bends to.
  The drumstick is attached to the tip of the manipulator with a soft sponge, which makes its control more difficult with conventional methods, so we can verify the benefit of the learning control system.
  The entire experimental setup is covered by a black curtain to make it easier to binarize the image.
}%
{%
  まず, 太鼓を叩くというタスクにおける具体的な実験セットアップを\figref{figure:experimental-setup}に示す.
  マニピュレータの自由度は2であり, サーボモータとしてDynamixel Motor (XM430-W210-R)を使用している.
  RGB SensorとしてはD435を用いており, マニピュレータの横には音を取るためのマイクを設置している.
  リンクはプラスチック製の薄いまな板に穴を開けて使用しており, 非常に柔軟な構造をしている.
  柔軟な構造は, 太鼓を叩くというタスクにおいて衝撃を吸収できるという意味で優れいる反面, 曲がり具合によって綺麗に音が出るかやタイミングが異なるため, 制御は難しい.
  アームの先端には太鼓を叩くバチがついており, モータとバチの間には柔らかなスポンジを挟むことでより制御を難しくし, 学習型制御の利点がわかるようにしている.
  実験セットアップ全体は暗幕により覆われ, 二値化処理がしやすいようにしている.
}%

\subsection{Training of DTXNET}
\begin{figure}[t]
  \centering
  \includegraphics[width=0.9\columnwidth]{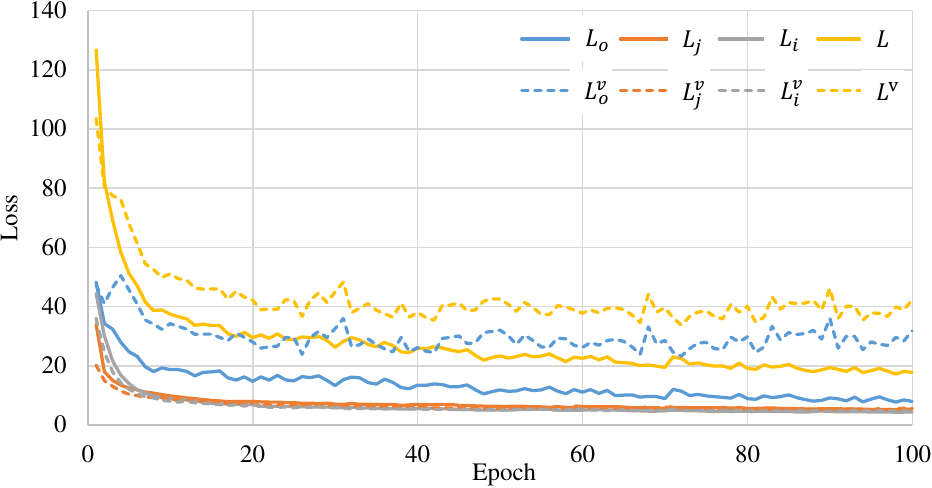}
  \caption{Loss transition when using the model of Type $3^+$ at training phase. $L_o$, $L_j$, $L_i$, and $L$ are the losses of task output, joint state, image, and their sum, respectively. $L^v_o$, $L^v_j$, $L^v_i$, and $L^v$ are the losses at validation phase.}
  \label{figure:train-type3}
  \vspace{-3.0ex}
\end{figure}

\begin{figure}[t]
  \centering
  \includegraphics[width=0.9\columnwidth]{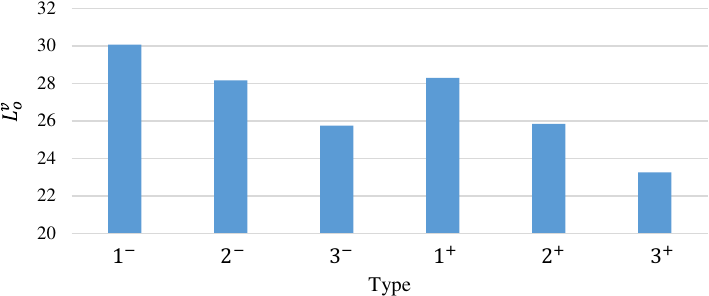}
  \caption{Comparison of output loss $L^v_o$ at validation phase, regarding all types.}
  \label{figure:train-comparison}
  \vspace{-3.0ex}
\end{figure}

\begin{figure}[t]
  \centering
  \includegraphics[width=1.0\columnwidth]{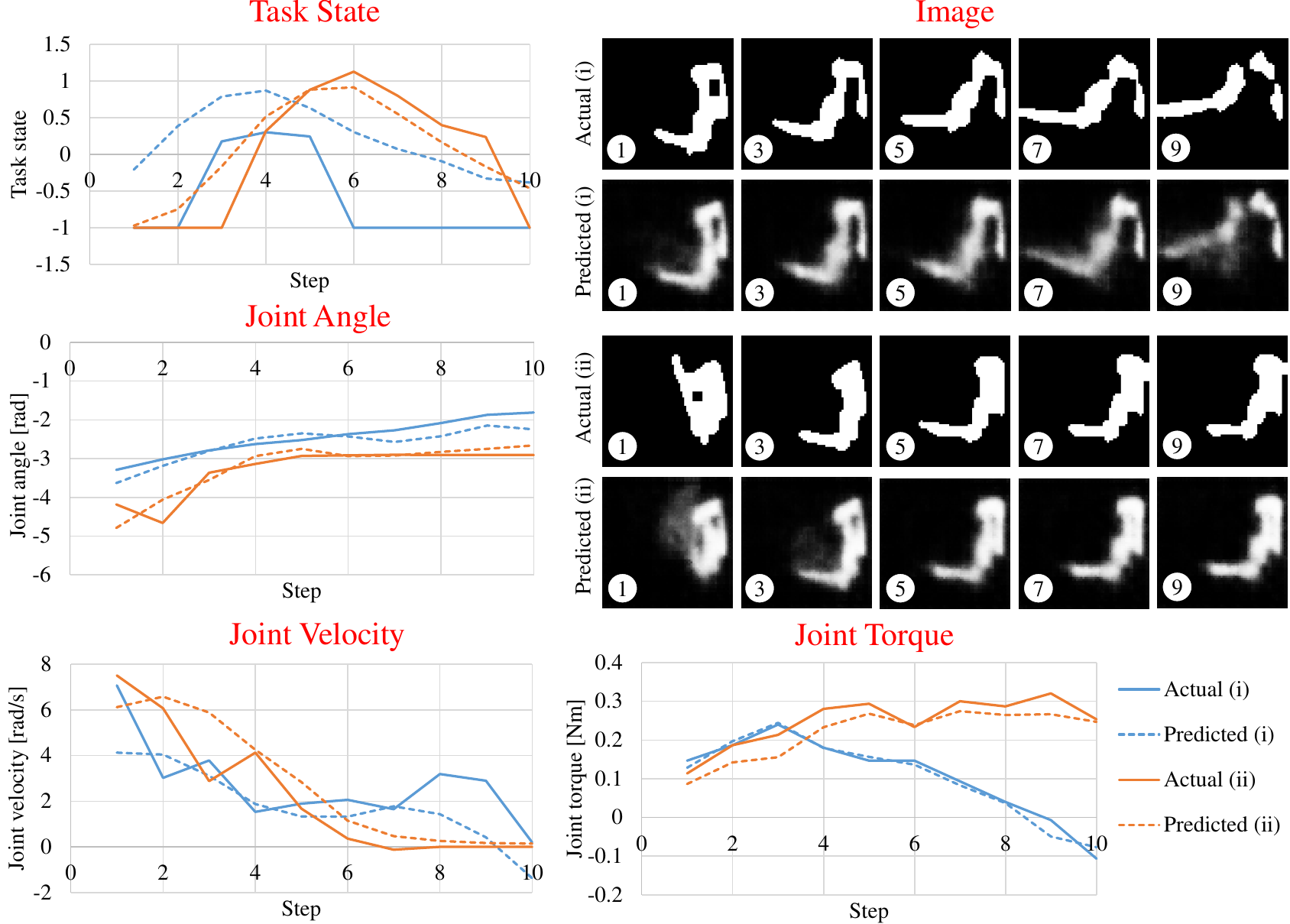}
  \caption{Inference results of task state, joint angle, joint velocity, joint torque, and image, when using the model of Type $3^+$.}
  \label{figure:inference-type3}
  \vspace{-3.0ex}
\end{figure}

\switchlanguage%
{%
  We conducted experiments regarding how well DTXNET can predict task state and robot state.

  First, we gave a random torque to the robot and obtained motion data for 5 minutes.
  In this process, we determined the maximum change of joint torque $du_{max} = 0.1$ [Nm] in one step, and sent the sum of the previous target torque and random value in $[-du_{max},\;du_{max}]$ to the robot.
  The control frequency is 15 Hz, so we obtained 4500 steps of data in 5 min.

  Of these data, we used 80$\%$ as training data and the rest as test data, and trained DTXNET for 100 epoch.
  The transition of losses $L_i, L_j, L_o, L$ and validation losses $L^v_i, L^v_j, L^v_o, L^v$ when using the model of Type $3^{+}$, are shown in \figref{figure:train-type3}.
  We can see that $L_i, L_j$ and $L^v_i, L^v_j$ decreased in the same way, but the prediction of $L^v_o$ was difficult.

  Because predicting task state is the most important in achieving the task, we use the model with the lowest $L^v_o$ in all epoch at control phase.
  We show the lowest $L^v_o$ of each type, in \figref{figure:train-comparison}.
  Regarding the $L^v_o$, we can see simple relationships between types: $1^- > 2^- > 3^-, 1^+ > 2^+ > 3^+, 1^- > 1^+, 2^- > 2^+, 3^- > 3^+$.

  \figref{figure:inference-type3} shows examples of inference results when using the model of Type $3^+$ with the lowest $L^v_o$ in all types.
  These results were obtained in the two situations named (i) and (ii).
  The graphs in \figref{figure:inference-type3} show the predicted values from DTXNET (Predicted) and actual values (Actual) regarding task state, joint angle, joint velocity, joint torque, and image, respectively.
  We expanded DTXNET and obtained the predicted values until 10 steps ahead.
  We can see joint angle, joint velocity, joint torque, and image were predicted almost correctly.
  Regarding task state, the inference of (ii) was good, but the inference of (i) had some errors, although the shapes of the graphs were almost the same.
}%
{%
  DTXNETによってどの程度タスクの状態, またはロボットの身体状態を予測できるかについて実験する.

  まずはrandomな関節トルクを与え, 5分間の動作データを取得する.
  関節トルクの変化幅の最大値$du_{max} = 0.1$ [Nm]を決め, 現在指令トルクに$-du_{max}$から$du_{max}$までのランダムな値を足しこみ, 実機に送る.
  関節角度の最小値$\theta_{min}$と最大値$\theta_{max}$を決めておき, その範囲を超えた場合はそれを戻す方向に$du_{max}$だけトルクを足しこみ実機に送る.
  動作周期は15 Hzなため, 5 minで4500ステップ分のデータを得ることができる.

  これらデータから80$\%$をtrain, 20$\%$をtestとして用い, 100 epoch学習を行う.
  例として, その際のtype $3^+$における$L_i, L_j, L_o, L$とvalidationの際のそれぞれのlossである$L^v_i, L^v_j, L^v_o, L^v$の変化を\figref{figure:train-type3}に示す.
  train, testともに$L_i, L_j$はしっかり下がるものの, 特に$L^v_o$の推論は難しいことがわかる.
  本研究では, 最終的にタスク状態の予測ができることが重要であるため, 全epochの中で$L^v_o$が最も小さいときのモデルをControl phaseでは使用する.

  \figref{figure:train-comparison}に全typeにおける$L^v_o$の最小値を示す.
  $L^v_o$は, typeにおいて$1^- > 2^- > 3^-, 1^+ > 2^+ > 3^+, 1^- > 1^+, 2^- > 2^+, 3^- > 3^+$という綺麗な関係があることがわかる.

  最も良いタスクの予測性能を持つtype $3^+$において, task状態, joint state, imageをどの程度推論できるかの例を\figref{figure:inference-type3}に示す.
  Sample (1, 2)の２つのサンプルを用意し, それぞれtask状態, joint angle, joint velocity, joint torque, imageに関して実際の値ActualとDTXNETによる予測Predictedを示している.
  DTXNETをステップ数は10まで展開している.
  joint angle, joint velocity, joint torque, imageは概ね予測出来ていることがわかる.
  Outputを予測するのは難しく, Sample 1は予測の形こそあっているもの, 大きさには誤差が生じしていることもわかる.
}%

\subsection{Wadaiko drumming with DTXNET}
\begin{figure}[t]
  \centering
  \includegraphics[width=0.9\columnwidth]{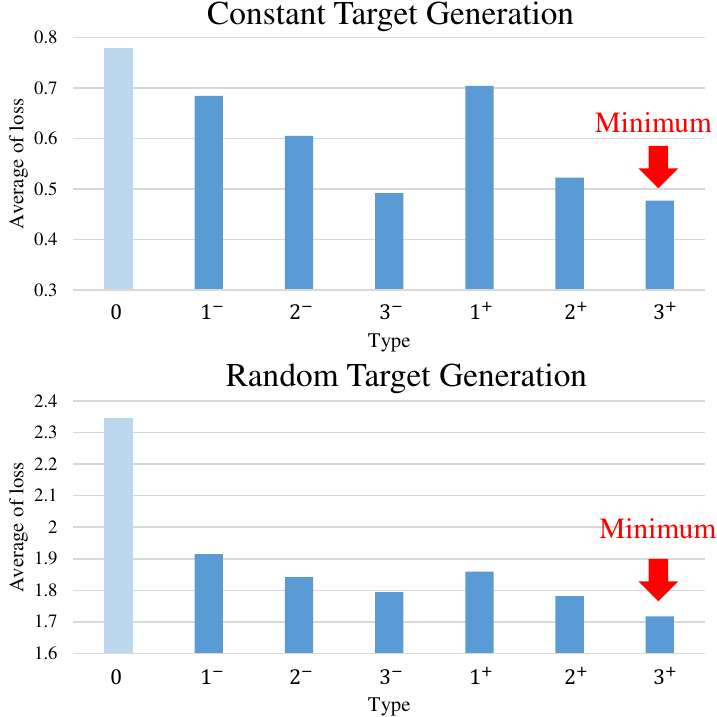}
  \caption{Comparison of the average of $L^{control}_o$ regarding Type $0,\;1^-,\;2^-,\;3^-,\;1^+, \;2^+$, and $3^+$, and regarding Constant and Random Target Generation.}
  \label{figure:average-loss}
  \vspace{-3.0ex}
\end{figure}

\begin{figure*}[t]
  \centering
  \includegraphics[width=2.0\columnwidth]{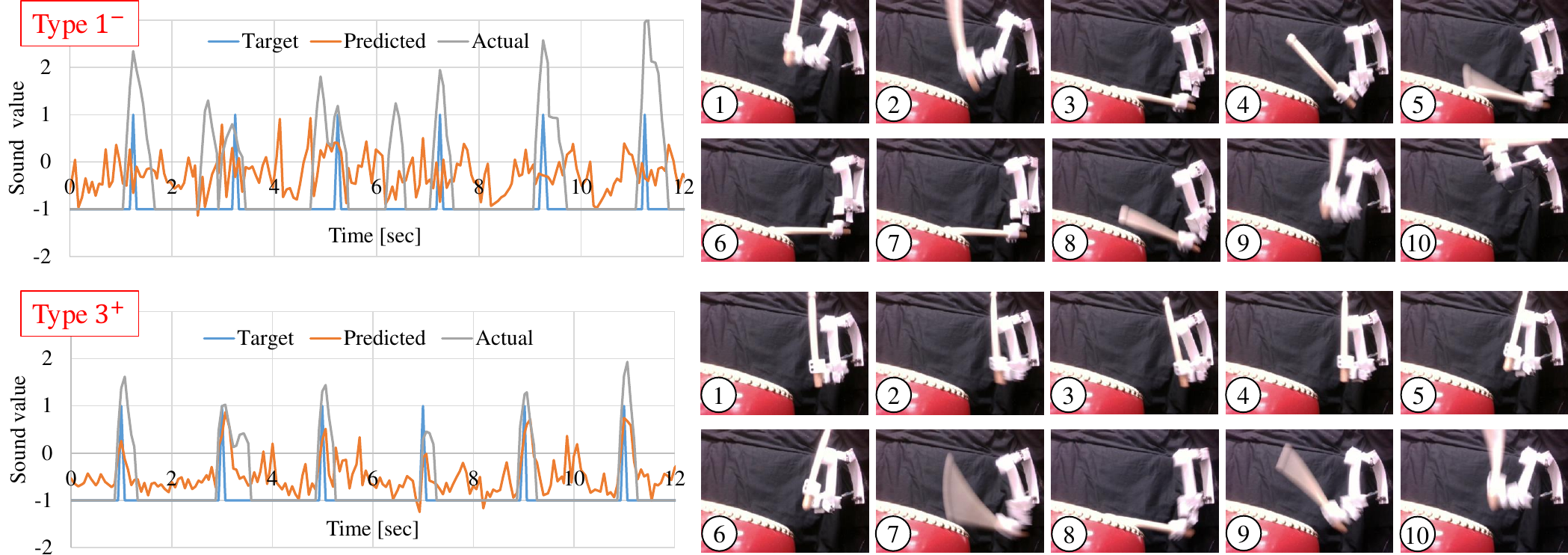}
  \caption{Comparison of target, predicted, and actual sound when using the models of Type $1^-$ and $3^+$, regarding Constant Target Generation.}
  \label{figure:control-comparison}
  \vspace{-3.0ex}
\end{figure*}

\switchlanguage%
{%
  By using DTXNET trained in the previous section, we conducted experiments with a Wadaiko drumming task.
  As evaluation, we used $L^{control}_{o}$ (\equref{eq:loss}) with the actual sound substituted for $o_{predicted}$.
  We executed the proposed real-time control using DTXNET for 1 min, and evaluated the average of $L^{control}_{o}$.
  Regarding the six types of $1^-, 2^-, 3^-, 1^+, 2^+, 3^+$, and Type 0 of random control without any optimization, we conducted experiments with Constant Target Generation, which sends $o_{target}=1$ every 2 sec, and with Random Target Generation, which sends a random value of $o_{target}$ in $[0,\;2]$ with 10$\%$ probability every step.
  We show the results in \figref{figure:average-loss}.
  We can see that controls of all types with DTXNET are better than Type 0.
  Also, we can see the same relationship with the result of training phase, and Type $3^+$ was the best controller.

  To understand the actual movements, regarding Type $1^-$ and Type $3^+$ at Constant Target Generation, we show the comparison of target, predicted, and actual sound volume in the left figure of \figref{figure:control-comparison}, and their movements in the right figure of \figref{figure:control-comparison}.
  The right figure is the sequence of 10 images from 4 to 5.5 sec.
  Regarding Type $1^-$, the predicted task state vibrated to a great extent, the actual task state was mostly larger than the target task state, and the actual sound was sometimes produced at a timing different from the target.
  As the right image sequence shows, the timing at which the robot should beat was only at \textcircled{8}, but in Type $1^-$, the robot beat several times before the target timing.
  On the other hand, regarding $3^+$, the predicted and actual values were close to the target value, and there were few unnecessary movements in right figure of \figref{figure:control-comparison}.
}%
{%
  前章で学習されたDTXNETを用いて, 太鼓を叩くタスクを行う.
  この際の評価は, \equref{eq:loss}の$o_{predicted}$に実際に出た音を代入した値$L^{control}_{o}$を用いて行う.
  それぞれのtypeにおいて, 提案したリアルタイム制御を1 min実行し, その際の$L^{control}_{o}$の平均により評価を行う.
  全6つのtype $1^-, 2^-, 3^-, 1^+, 2^+, 3^+$と一切の制御せずにランダムに動かすtype 0において, 大きさ1の指令を2秒間隔で送るConstant target generation, 0から2の大きさの指令を毎ステップ10$\%$の確率で指令するRandom target generationを行った.
  \figref{figure:average-loss}にその結果を示す.
  DTXNETを用いた全てのtypeにおいて, 制御なしのtype 0よりも指令値を実現できていることがわかる.
  また, Training phaseと同様の関係が見られ, type $3^+$が最も良い制御器を実現できている.

  動きの違いがわかりやすいよう, type $1^-$とtype $3^+$におけるConstant target generationでの音の指令値, 予測値, 実際値の比較を\figref{figure:control-comparison}の左図に, その際の動作画像を右図に示す.
  右図の様子は左図のグラフの4 secから5.5 secまで0.15 secごとに10フレームを取った際の動作画像列である.
  まず, type $1^-$とtype $3^+$の大きな違いとして, 前者は予測が大きく振動しており, 実際に出る音も指令値に比べて大きかったり, 指令がないタイミングで鳴っていたりしている.
  右図の画像列を見ればわかるように, 最終的に音を出すべきは\textcircled{8}の瞬間であるが, その前の段階で何度か太鼓を叩いてしまっていることがわかる.
  それに対してtype $3^+$では, 予測, 実際の値ともに指令値に近く, 右図の画像列からもわかるように動作の際に無駄な動きが少ない.
}%

% \subsection{Performance} \label{subsec:performance}
% \switchlanguage%
% {%
% }%
% {%
%   やるか微妙なところ.
% }%

\section{Discussion} \label{sec:discussion}
\switchlanguage%
{%
  In all experiments, the network with an input and output of Joint State and Image (Type $3^+$) was the best.
  This means that each Joint State and Image includes important information, which is not included in the other, to realize the task.
  Joint State has the current joint torque information, but it cannot be reconstructed from the current image and optical flow.
  In the same manner, to determine the posture of the flexible body is difficult from just Joint State.
  Also, by including robot states in the output of DTXNET, the network can obtain the losses other than those of sparse sound information, and so a more correct training can proceed.

  DTXNET can use variable $T$, and we can adjust the $T$ without retraining DTXNET.
  This is an advantage from EMD Net \cite{tanaka2018emd} and Dynamics-Net \cite{kawaharazuka2019dynamic} that predict the states after a fixed time step $T$.
  However, this can also be a problem, because DTXNET is expanded sequentially and needs more calculation time.
  Another superior point of DTXNET is that while it takes large computational cost to reconstruct robot states at training phase, it takes smaller computational cost at control phase, because there is no need to output robot states.
  In this study, although we can execute \equref{eq:opt} only once at each time step to reduce the computational cost, the control command is optimized $T_{control}$ times before sending it, and so it was enough to realize the task.

  In our experiments, we used Joint State and Image as robot states, sound volume as task state, and joint torque as control command.
  In actuality, DTXNET can handle joint angle, joint velocity, air mass flow, etc. as control command, using the same network structure.
  Also, DTXNET has potential to handle values other than sound, such as the position of the arm tip and the degree of object deformation, as task state.
  Regarding the soft robot with redundant sensors, DTXNET can handle air pressure, strain gauge, infrared sensor, etc., as robot state.
  In future works, we would like to consider various tasks by various flexible bodies with redundant sensors.
}%
{%
  まず, 本研究で行った実験結果について考察する.
  どの実験においても, Joint StateとImageを入力し, それらを出力にも加えるようなtype $3^+$のネットワークの性能が最も良かった.
  これは, Joint State, Imageそれぞれに, 一方にはないタスク実行に関する重要な情報が内包されているからであると考えられる.
  Joint Stateは現在の関節トルク情報を持つ一方, 画像とそのオプティカルフローからだけではそれを復元することは難しい.
  そして, Joint Stateの関節角度・角速度からでは, 柔軟マニピュレータの現在の姿勢を決めることは難しいということである.
  また, DTXNETの出力に身体状態を含めることで, 情報が疎なloss以外についてのlossが返るため, より正しい予測の進行がしやすくなると考えられる.

  次に, DTXNETの構造に関する考察を行う.
  EMD Net \cite{tanaka2018emd}やDynamics-Net \cite{kawaharazuka2019dynamic}では, 現在状態から固定されたステップ後の状態を予測していた.
  それに対してDTXNETは, $T$を可変にすることができ, 計算時間等から再学習なしにそれを調整することができる.
  これは同時に問題点ともなり, $T$を固定して単一の全結合層で表すモデルと比べ, sequentialにNetworkを展開するため実行時間は遅くなる.
  また, DTXNETの構造の優れている点として, Training phaseでは画像を復元するような計算量の多い処理を行うが, リアルタイム制御実行時はその部分に関する計算を必要としないため, 計算量を減らすことができるという点が挙げられる.
  本モデルは計算量的な観点から一度しか\equref{eq:opt}を実行できていないが, 制御入力は$T_{control}$ステップ分だけ毎回最適化された結果実機に送られるため, タスクを実現するのに十分な最適化のループを回すことができている.

  最後に, DTXNETの汎用性について議論する.
  本研究の実験では, DTXNETが考慮するロボット状態をJoint State, Image, タスク状態を音の大きさ, 制御入力を関節トルクとしていた.
  実際には, DTXNETは同じ構造で制御入力として関節角度や関節速度, 空気の流入量等を扱うことができる.
  また, タスク状態も手先位置や物体の変形具合等, 音以外の様々な情報を扱うことができる.
  冗長なセンサを持つソフトロボットにおいては, 空気の圧力やひずみゲージ, 赤外線センサ等をロボット状態として扱うこともできる.
  今後は, 冗長なセンサを有する様々な柔軟身体を扱ったタスク実行についても考察していきたい.
}%

\section{CONCLUSION} \label{sec:conclusion}
\switchlanguage%
{%
  In this study, we proposed a learning control method to realize dynamic tasks by a flexible manipulator.
  By constructing DTXNET with LSTM, we can predict the task state variable time steps ahead, and realize dynamic control of a flexible manipulator using image information.
  We can propose six types of configurations of DTXNET, depending on if the network predicts robot states or not, and if the network uses actuator or image information as robot states.
  In these configurations, by using both the actuator and image information and adding the limitation of DTXNET to predict robot states, we can control the flexible manipulator more precisely.

  In future works, we would like to realize various tasks by various manipulators, and develop a system to realize multiple tasks continuously using a flexible manipulator.
}%
{%
  本研究では, 柔軟なマニピュレータによる疎な状態を扱う動的タスクを, 学習的に実現する手法を提案した.
  LSTMを用いたDTXNETを構築することで予見するステップを可変にでき, 画像を用いることで柔軟なマニピュレータの正確な制御を実現している.
  DTXNETにはロボットの観測状態を予測するかどうか, 観測状態としてアクチュエータ情報と同時に画像情報を用いるかどうかから6種類の構造が提案できる.
  その中でも画像とアクチュエータ情報の両者を用い, DTXNETがロボットの観測状態も予測できるような制限を加えることで, より正確な制御を実現できることがわかった.

  今後は, より様々な柔軟マニピュレータにより多くのタスクを実現すること, また, 柔軟なマニピュレータにより一つのタスクではなく複数のタスクをcontinuousに続けていけるようなシステムを開発していきたい.
}%

{
  %\footnotesize
  %\small
  %\bibliographystyle{junsrt}
  \bibliographystyle{IEEEtran}
  \bibliography{main}

\begin{thebibliography}{10}
\providecommand{\url}[1]{#1}
\csname url@rmstyle\endcsname
\providecommand{\newblock}{\relax}
\providecommand{\bibinfo}[2]{#2}
\providecommand\BIBentrySTDinterwordspacing{\spaceskip=0pt\relax}
\providecommand\BIBentryALTinterwordstretchfactor{4}
\providecommand\BIBentryALTinterwordspacing{\spaceskip=\fontdimen2\font plus
\BIBentryALTinterwordstretchfactor\fontdimen3\font minus
  \fontdimen4\font\relax}
\providecommand\BIBforeignlanguage[2]{{%
\expandafter\ifx\csname l@#1\endcsname\relax
\typeout{** WARNING: IEEEtran.bst: No hyphenation pattern has been}%
\typeout{** loaded for the language `#1'. Using the pattern for}%
\typeout{** the default language instead.}%
\else
\language=\csname l@#1\endcsname
\fi
#2}}

\bibitem{hirai1998asimo}
K.~Hirai, M.~Hirose, Y.~Haikawa, and T.~Takenaka, ``{The Development of Honda
  Humanoid Robot},'' in \emph{Proceedings of the 1998 IEEE International
  Conference on Robotics and Automation}, 1998, pp. 1321--1326.

\bibitem{kim2013softrobotics}
S.~Kim, C.~Laschi, and B.~Trimmer, ``{Soft robotics: a bioinspired evolution in
  robotics},'' \emph{Trends in Biotechnology}, vol.~31, no.~5, pp. 287--294,
  2013.

\bibitem{lee2017softrobotics}
C.~Lee, M.~Kim, Y.~J. Kim, N.~Hong, S.~Ryu, H.~J. Kim, and S.~Kim, ``{Soft
  robot review},'' \emph{International Journal of Control, Automation and
  Systems}, vol.~15, no.~1, pp. 3--15, 2017.

\bibitem{escande2014qp}
A.~Escande, N.~Mansard, and P.-B. Wieber, ``{Hierarchical quadratic
  programming: Fast online humanoid-robot motion generation},'' \emph{The
  International Journal of Robotics Research}, vol.~33, no.~7, pp. 1006--1028,
  2014.

\bibitem{wieber2006mpc}
P.~Wieber, ``{Trajectory Free Linear Model Predictive Control for Stable
  Walking in the Presence of Strong Perturbations},'' in \emph{Proceedings of
  the 2006 IEEE-RAS International Conference on Humanoid Robots}, 2006, pp.
  137--142.

\bibitem{kiang2015flexible}
C.~T. Kiang, A.~Spowage, and C.~K. Yoong, ``{Review of Control and Sensor
  System of Flexible Manipulator},'' \emph{Journal of Intelligent {\&} Robotic
  Systems}, vol.~77, no.~1, pp. 187--213, 2015.

\bibitem{book1984flexible}
W.~J. Book, ``{Recursive Lagrangian Dynamics of Flexible Manipulator Arms},''
  \emph{The International Journal of Robotics Research}, vol.~3, no.~3, pp.
  87--101, 1984.

\bibitem{kotnik1988acceleration}
P.~T. Kotnik, S.~Yurkovich, and {\"U}.~{\"O}zg{\"u}ner, ``{Acceleration
  feedback for control of a flexible manipulator arm},'' \emph{Journal of
  Robotic Systems}, vol.~5, no.~3, pp. 181--196, 1988.

\bibitem{moudgal1995flexible}
V.~G. Moudgal, W.~A. Kwong, K.~M. Passino, and S.~Yurkovich, ``{Fuzzy learning
  control for a flexible-link robot},'' \emph{IEEE Transactions on Fuzzy
  Systems}, vol.~3, no.~2, pp. 199--210, 1995.

\bibitem{pradhan2012flexible}
S.~K. Pradhan and B.~Subudhi, ``{Real-Time Adaptive Control of a Flexible
  Manipulator Using Reinforcement Learning},'' \emph{IEEE Transactions on
  Automation Science and Engineering}, vol.~9, no.~2, pp. 237--249, 2012.

\bibitem{su2001flexible}
Z.~Su and K.~Khorasani, ``{A neural-network-based controller for a single-link
  flexible manipulator using the inverse dynamics approach},'' \emph{IEEE
  Transactions on Industrial Electronics}, vol.~48, no.~6, pp. 1074--1086,
  2001.

\bibitem{inaba1987rope}
M.~Inaba and H.~Inoue, ``{Rope handling by a robot with visual feedback},''
  \emph{Advanced Robotics}, vol.~2, no.~1, pp. 39--54, 1987.

\bibitem{yamakawa2011folding}
Y.~Yamakawa, A.~Namiki, and M.~Ishikawa, ``{Motion planning for dynamic folding
  of a cloth with two high-speed robot hands and two high-speed sliders},'' in
  \emph{Proceedings of the 2011 IEEE International Conference on Robotics and
  Automation}, 2011, pp. 5486--5491.

\bibitem{tanaka2018emd}
D.~Tanaka, S.~Arnold, and K.~Yamazaki, ``{EMD Net: An Encode-Manipulate-Decode
  Network for Cloth Manipulation},'' \emph{IEEE Robotics and Automation
  Letters}, vol.~3, no.~3, pp. 1771--1778, 2018.

\bibitem{kawaharazuka2019dynamic}
K.~Kawaharazuka, T.~Ogawa, J.~Tamura, and C.~Nabeshima, ``{Dynamic Manipulation
  of Flexible Objects with Torque Sequence Using a Deep Neural Network},'' in
  \emph{Proceedings of the 2019 IEEE International Conference on Robotics and
  Automation}, 2019, pp. 2139--2145.

\bibitem{yang2017repeatable}
P.~Yang, K.~Sasaki, K.~Suzuki, K.~Kase, S.~Sugano, and T.~Ogata, ``{Repeatable
  Folding Task by Humanoid Robot Worker Using Deep Learning},'' \emph{IEEE
  Robotics and Automation Letters}, vol.~2, no.~2, pp. 397--403, 2017.

\bibitem{lee2015learning}
A.~X. Lee, H.~Lu, A.~Gupta, S.~Levine, and P.~Abbeel, ``{Learning force-based
  manipulation of deformable objects from multiple demonstrations},'' in
  \emph{Proceedings of the 2015 IEEE International Conference on Robotics and
  Automation}, 2015, pp. 177--184.

\bibitem{wu2018flexible}
Y.~Wu, K.~Takahashi, H.~Yamada, K.~KIM, S.~Murata, S.~Sugano, and T.~Ogata,
  ``{Dynamic Motion Generation by Flexible-Joint Robot based on Deep Learning
  using Images},'' in \emph{Proceeding of the 8th Joint IEEE International
  Conference on Development and Learning on Epigenetic Robotics}, 2018.

\bibitem{hochreiter1997lstm}
S.~Hochreiter and J.~Schmidhuber, ``{Long short-term memory},'' \emph{Neural
  computation}, vol.~9, no.~8, pp. 1735--1780, 1997.

\bibitem{rumelhart1986bptt}
D.~E. Rumelhart, G.~E. Hinton, and R.~J. Williams, ``{Learning Internal
  Representations by Error Propagation},'' in \emph{Parallel Distributed
  Processing: Explorations in the Microstructure of Cognition, Vol. 1}, D.~E.
  Rumelhart, J.~L. McClelland, and C.~PDP Research~Group, Eds.\hskip 1em plus
  0.5em minus 0.4em\relax Cambridge, MA, USA: MIT Press, 1986, pp. 318--362.

\bibitem{ioffe2015batchnorm}
S.~Ioffe and C.~Szegedy, ``{Batch Normalization: Accelerating Deep Network
  Training by Reducing Internal Covariate Shift},'' in \emph{Proceedings of the
  32nd International Conference on Machine Learning}, 2015, pp. 448--456.

\bibitem{tokui2015chainer}
S.~Tokui, K.~Oono, S.~Hido, and J.~Clayton, ``{Chainer: a Next-Generation Open
  Source Framework for Deep Learning},'' in \emph{Proceedings of Workshop on
  Machine Learning Systems in The 29th Annual Conference on Neural Information
  Processing Systems}, 2015.

\bibitem{kingma2015adam}
D.~P. Kingma and J.~Ba, ``{Adam: A Method for Stochastic Optimization},'' in
  \emph{Proceedings of the 3rd International Conference on Learning
  Representations}, 2015, pp. 1--15.

\end{thebibliography}
}

\end{document}